# Adversarial Attacks on Brain-Inspired Hyperdimensional Computing-Based Classifiers

Fangfang Yang
UC Riverside
fyang050@ucr.edu

Shaolei Ren
UC Riverside
sren@ece.ucr.edu

## ABSTRACT

Being an emerging class of in-memory computing architecture, brain-inspired hyperdimensional computing (HDC) mimics brain cognition and leverages random hypervectors (i.e., vectors with a dimensionality of thousands or even more) to represent features and to perform classification tasks. The unique hypervector representation enables HDC classifiers to exhibit high energy efficiency, low inference latency and strong robustness against hardware-induced bit errors. Consequently, they have been increasingly recognized as an appealing alternative to or even replacement of traditional deep neural networks (DNNs) for local on-device classification, especially on low-power Internet of Things devices. Nonetheless, unlike their DNN counterparts, state-of-the-art designs for HDC classifiers are mostly security-oblivious, casting doubt on their safety and immunity to adversarial inputs. In this paper, we study for the first time adversarial attacks on HDC classifiers and highlight that HDC classifiers can be vulnerable to even minimally-perturbed adversarial samples. Concretely, using handwritten digit classification as an example, we construct a HDC classifier and formulate a grey-box attack problem, where an attacker's goal is to mislead the target HDC classifier to produce erroneous prediction labels while keeping the amount of added perturbation noise as little as possible. Then, we propose a modified genetic algorithm to generate adversarial samples within a reasonably small number of queries, and further apply critical gene crossover and perturbation adjustment to limit the amount of perturbation noise. Our results show that adversarial samples generated by our algorithm can successfully mislead the HDC classifier to produce wrong prediction labels with a high probability (i.e., 78% when the HDC classifier uses a fixed majority rule for decision). Finally, we also present two defense strategies — adversarial training and retraining — to strengthen the security of HDC classifiers.

## 1 INTRODUCTION

With the exploding rise of the Internet of Things (IoT) and edge computing, there comes an enormous amount of data that is being continuously generated at countless devices all over the world [1]. Nonetheless, due to various constraints such as timing, privacy concerns and lack of sufficient bandwidth, moving all the data to central clouds for processing is simply out of question. Consequently, the need of on-device machine learning inference (e.g., running image classification on smart cameras powered by deep neural networks or DNNs) to extract actionable information from locally generated data has been quickly surging [2, 3].

To run complex DNNs for inference on resource-constrained devices, numerous model compression and neural architecture search (NAS) techniques have been proposed, thereby reducing the DNN model size and inference latency [3–10]. Thanks to these efforts, DNN inference on resource-constrained edge devices has become a reality, with state-of-the-art compressed DNN models being able to deliver near real-time image classification on modern mobile devices [3, 6]. Despite these encouraging results, however, the room left for further reducing the inference latency and energy consumption is vanishing, which adds significant challenges to ubiquitously fit DNN models into a full spectrum of edge devices such as wearables and low-cost IoT devices. This is due in part to the inherent limitation of DNN model and architecture that require overly intensive mathematical operation and computing beyond the capability of many edge devices.

In very recent years, brain-inspired hyperdimensional computing (HDC) has emerged as an ultra-lightweight *classification* framework and architecture [11–14]. Specifically, HDC exploits the key principle that human brain "computes" based on certain patterns formed by a large number of neurons, without being directly associated with numbers [13]. Thus, instead of computing with numbers like in today's DNNs, a HDC classifier mimics the way brain cognition works by representing information/features using a hypervector with binary elements in a very high-dimensional space (e.g., with a dimensionality of $D = 10^4$ or more), where there are numerous hypervectors that are almost certainly orthogonal to each other [12]. Through training based on simple operation such as superposition and permutation (details in Section 2), a set of binary hypervectors representing different prediction classes can be constructed and stored in an associative memory. For testing/inference, a query sample is also mapped to a hypervector, which is then compared against pre-trained class hypervectors in the associative memory to produce a prediction label based on a distance metric (e.g., typically Hamming distance) [11, 13].

Unlike standard DNN-based classifiers that require non-linear multi-layer perception, classification based on HDC is inherently "in-memory" due to their binarized hypervectors and can be performed using basic logical operations like XOR without the need of sophisticated computation [13, 15]. As a result, compared to DNN-based classifiers, HDC classifiers offer several key advantages, including extremely high energy efficiency, low latency, and strong robustness against hardware-induced component failures [12, 13]. For example, recent studies have shown that the energy consumption and inference latency of HDC classifiers are lower by orders of magnitude than their DNN counterparts, yet achieving a reasonable inference accuracy [14, 16–21]. Moreover, in HDC, binary elements are independently and identically distributed (i.i.d.) and not a single element carries more information than others in a hypervector with a large dimensionality. Therefore, a small number of memory-induced random bit errors in the associative memory can barely affect the classification result, making HDC classifiers less prone to hardware errors [13, 22]. Additionally, the training process for HDC classifiers is also much simpler than for DNNs, using only basic logic operators without complex optimization techniques [12, 13, 15]. In Section 3, we will provide an example of building a HDC classifier for handwritten digit classification.

Consequently, HDC classifiers have been increasingly recognized as an alternative to or even replacement of DNNs for local on-device classification, especially on low-power devices [12, 13, 15]. For example, the quickly expanding list of applications building on HDC classifiers have already included language classification [23], image classification [11, 24], emotion recognition based on physiological Signals [25], distributed fault isolation in power plants [26], gesture recognition for wearable devices [21], and seizure onset detection and identification of ictogenic brain regions [27]. More recently, HDC-based learning has also been integrated with visual perception modules of safety-critical robots for real-time navigation [28].

Nonetheless, despite the aforementioned appealing advantages compared to DNNs, the security aspect of emerging HDC classifiers has not been clearly understood. It is well-known that traditional DNNs can be highly vulnerable to adversarial samples that look almost identical to correctly-classified benign samples by human perception but are still misclassified, thereby driving a surging interest in safeguarding DNNs against adversarial inputs [29–33]. By contrast, the existing studies on HDC classifiers have been predominantly focused on reducing energy consumption and latency or further improving the HDC architecture [14, 16–21, 34], while the security aspect remains untouched. This can raise serious concerns with the safety of HDC classifiers and limit their wider adoption, especially in mission-critical applications such as robot navigation and health monitoring [27, 28].

**Contribution.** In this paper, we make a first-of-its-kind effort to investigate the potential vulnerability of emerging HDC classifiers. We highlight that state-of-the-art designs of HDC classifiers [17, 18] are security oblivious and their robustness against hardware-induced bit errors [11, 12, 22] does *not* imply robustness against adversarial inputs. Using handwritten digit classification as an example, we demonstrate that HDC classifiers suffer from vulnerability to minimally-perturbed adversarial samples. In Fig. 1, we show an example of adversarial attacks. While the adversarial image looks almost identical to the benign one with differences in only a few pixels, the HDC classifier is fooled and can only predict the adversarial image as digit "2".

More concretely, we build a HDC classifier for handwritten digit classification based on the MNIST dataset [35], following the widely-used training approach in HDC [11, 12]. Then, we consider a threat model in which an attacker can launch grey-box attacks by repeatedly sending perturbed images to the HDC classifier and receiving the Hamming distance output as well as the prediction label from the classifier. We formulate this problem as regularized optimization, where the attacker's goal is to mislead the target HDC classifier to produce wrong prediction labels (i.e., non-targeted attack) while keeping the amount of added perturbation noise as little as possible.

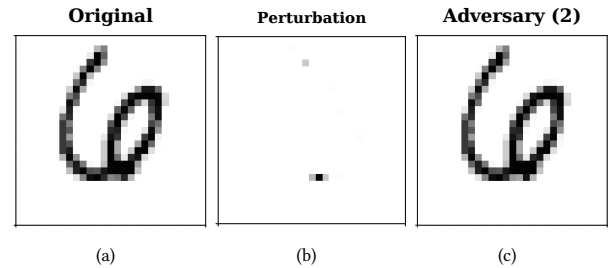

**Figure 1: An adversarial image generated by our modified genetic algorithm** GA-CGC-PA. **(a) Original benign image of digit "6". (b) Added perturbation noise, whose value is re-scaled to the maximum for better visualization and would otherwise be nearly invisible. (c) Adversarial image that looks almost identical to the benign image but is misclassified as "2".**

A key challenge for launching successful adversarial attacks is that the attacker does not know how the target HDC classifier represents features in a hyperdimensional space and instead only receives the Hamming distance output as well as the prediction label. In addition, the way that the target HDC classifier maps inputs to hypervectors is highly non-differentiable and thus, gradient-based approaches for grey-box attacks on standard DNNs do not apply [33]. To address these challenges, we propose a modified genetic algorithm, called Genetic Algorithm with Critical Gene Crossover and Perturbation Adjustment (GA-CGC-PA). Specifically, GA-CGC-PA only modifies critical genes (i.e., selected important pixels that are more relevant to classification) and iteratively searches for the best candidate adversarial image. GA-CGC-PA also applies perturbation adjustment to further reduce the amount of perturbation noise added to the original benign image.

Our evaluation results demonstrate that, for the majority of benign images under our test, the attacker can add a reasonably small amount of perturbation noise (i.e., as few as 9 pixels) and create adversarial images within a limited number of iterations, successfully misleading the target HDC classifier to a wrong prediction label. This presents a significant threat to HDC classifiers and requires appropriate defense strategies. To strengthen the security of HDC classifiers, we present two simple defense strategies to degrade attack performance in terms of the attack success rate (ASR): one based on the idea of adversarial training (i.e., including adversarial samples into the training dataset) [15, 33], and the other one based on the idea of retraining and ensemble learning [36].

In summary, we propose GA-CGC-PA to generate adversarial images that look similar to the original benign ones by human perception but are misclassified by HDC classifiers. We also discuss a set of possible defense mechanisms to enhance the security of HDC classifiers, making them more applicable in future safety-critical applications. While our threat model and attacks share similarities with conventional adversarial machine learning on DNNs [37, 38], we use a Hamming distance-related objective function in Eqn. (2), which is specifically tailored to HDC classifiers. Most importantly, the key novelty of our study is that it is the first to investigate the potential vulnerability of emerging brain-inspired HDC classifiers.

## 2 PRELIMINARIES ON HDC CLASSIFIERS

HDC classifiers have been found in an increasingly larger set of applications, including image classification [11, 24], seizure onset detection [27], fault isolation in power plants [26], robot navigation [28], among many others. An important property of HDC is that each hypervector is a pseudorandom $D$-dimensional vector taken by default from $\{-1, 1\}^D$ containing i.i.d. binary elements that can be either 1 or $-1$ [11]. Then, HDC leverages distributed holographic representation to project information onto a hyperdimensional space. Given two hypervectors, Hamming distance is commonly used as a distance metric to measure their similarity. Formally, Hamming distance is defined as the number of distinct binary elements between two hypervectors. If there exist $D/2$ different elements, then the inner product of two hypervectors is zero and the two hypervectors are considered orthogonal. For the convenience of presentation, Hamming distance is often normalized with respect to the dimensionality $D$ in HDC. Thus, two orthogonal hypervectors have a (normalized) Hamming distance of 0.5.

### 2.1 Random Indexing

A HDC classifier projects data onto a hyperdimensional space via random indexing. Specifically, a random hypervector in $\{-1, 1\}^D$ has an almost equal number of randomly placed -1 and 1. The almost-certain orthogonality due to the large dimensionality of $D$ demonstrates that any two randomly chosen hypervectors are orthogonal or quasi-orthogonal with an extremely high likelihood [11–13]. In a hyperdimensional space, there are enormous hypervectors that are orthogonal to each other. Such uncorrelated hypervectors can be used to represent various types of information or features of an object, such as 26 letters in the alphabet set. The hypervectors representing the basic features are called *basis* hypervectors, which remain unchanged in an application once randomly chosen. Moreover, through a set of well-defined operators performed over basis hypervectors, an object can then be encoded into a new hypervector for further classification [13, 22, 39].

### 2.2 Multiply-Add-Permute Operation

The most widely-used HDC operation is Multiply-Add-Permute (MAP), which specifies three fundamental operators: binding (multiplication), superposition (addition), and permutation.

**Binding (Multiplication).** Given two hypervectors $HV_1$ and $HV_2$, binding operation performs element-wise multiplication, denoted as $HV_1 \otimes HV_2$. The operation is used to represent the association of related hypervectors. The resulting hypervector of binding is orthogonal to both of its constituents [13].

**Superposition (Addition).** Like element-wise multiplication in binding operation, superposition of $HV_1, \cdots, HV_M$ is an element-wise addition of hypervectors denoted as $HV_1 \oplus \cdots \oplus HV_M$. Superposition aims to generate a sum hypervector $HV'$, which can represent a set of operand hypervectors and aggregate information conveyed by them. According to Hebbian Learning, after superposition, any of the constituents is more similar to $HV'$ than a randomly generated hypervector [40, 41].

An element-wise majority rule (MR) is routinely adopted after superposition, ensuring that the resulting hypervector still lies in the hyperdimensional space $\{-1, 1\}^D$. Concretely, if the component

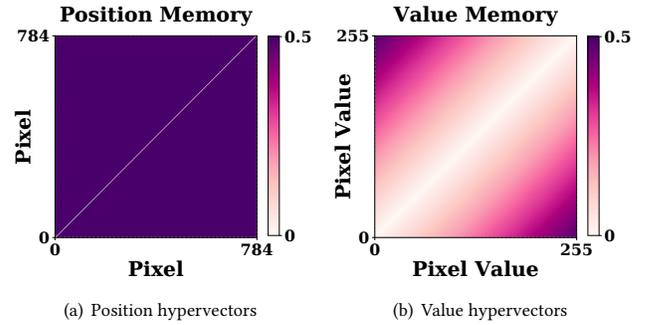

**(a)** Position hypervectors    **(b)** Value hypervectors

**Figure 2: Hamming distances. (a) The normalized Hamming distance between each two different position hypervectors is 0.5. (b) The normalized Hamming distance between each pair of value hypervectors for 256 pixel values: The larger the pixel value difference, the larger the Hamming distance between their corresponding value hypervectors.**

value of the resultant after addition is positive (i.e., there are more 1s than $-1$s in superposition), it is converted to 1 and otherwise $-1$. In the even that the component value of the resultant is zero, it is randomly encoded to 1 or $-1$ with equal probabilities, which we also refer to as the random majority rule (RMR) [42]. Alternatively, we can also always assign 1 or $-1$ to the component value in such cases, which we refer to as the fix majority rule (FMR).

**Permutation.** The permutation operation generates a dissimilar hypervector by shuffling coordinates of the original hypervector in a pseudo-random manner. A hypervector $HV$ permuted $n$ times is denoted as $\rho^n(HV)$. Permutation is used to store and differentiate the sequence of elements. For example, the letter sequence *abc* can be distinguished from *bac* by permutation.

## 3 A HDC CLASSIFIER ON MNIST DATASET

In this section, we construct a HDC classifier on the MNIST dataset [35] for handwritten digit recognition, which will be the target application for describing our proposed adversarial attacks. Our design builds on the widely-used approach to constructing HDC classifiers [11, 18, 22]. Note that it is also an active research direction in the field to design HDC classifiers for more complex image classification tasks [11, 43].

In the MNIST dataset, each digit image has $28 \times 28$ pixels, with integer pixel values in the range of $[0, 255]$. During the training stage, the key goal of our HDC classifier is to project each of the digit images in training dataset with labels to a hypervector (called *sample* hypervector) and then construct a single hypervector for each class (called *class* hypervector). The class hypervectors are then stored in an associative memory for inference on testing images. During the testing stage, an unknown test sample is also encoded to a hypervector (called *query* hypervector), which is then compared in parallel with all the trained class hypervectors in terms of the Hamming distances. Finally, the classifier returns the label of the class hypervector that has the shortest Hamming distance to the query hypervector.

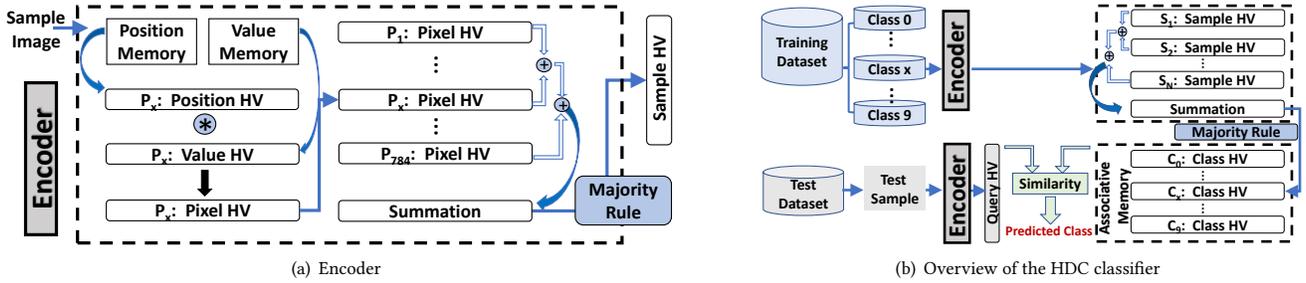

(a) Encoder

(b) Overview of the HDC classifier

**Figure 3: (a) The encoder in our HDC classifier encodes a digital image (called sample image) to a sample hypervector. (b) The overview of the HDC classifier. During the training stage, an associative memory storing class hypervectors is generated using the training dataset. Then, a test sample can be classified based on its similarity to class hypervectors.**

### 3.1 Mapping

In conventional machine learning, a handwritten digit image can be fully represented by its pixel values together with the corresponding pixel positions. In HDC, an image needs to be represented using hypervectors instead. Thus, we need to first map the information of pixel positions as well as the pixel values into a set of basis hypervectors, which will later be encoded into a single hypervector for image representation in a hyperdimensional space.

In order to convert an image's original representation to HDC representation, there are two main mapping techniques — orthogonal distributed mapping and distance preserving mapping [11].

Considering that there are $28 \times 28 = 784$ pixels in an image in MNIST, we employ orthogonal distributed mapping to encode the position information of each pixel. Concretely, we assign a random hypervector to each position (called *position* hypervector), which automatically ensures that the 784 position hypervectors are distinct and quasi-orthogonal to each other due to the hyperdimensionality. We store these position hypervectors in a look-up table, which is referred to as position memory. Fig. 2(a) shows the normalized Hamming distance between each pair of position hypervectors. We see that the normalized Hamming distance between any two different position hypervectors is almost 0.5, implying that the randomly chosen position hypervectors are quasi-orthogonal to each other [42].

Next, we map pixel values to hypervectors, which are called *value* hypervectors. Clearly, different pixel values are correlated. To preserve similarity of pixel values, we adopt the distance preserving mapping technique and create nearly similar value hypervectors to represent 256 pixel levels, since each pixel value in the MNIST dataset is stored as a 8-bit integer. Typically, the value hypervectors associated with the minimum and maximum pixel values are orthogonal (i.e. there are $\frac{D}{2}$ distinct elements between the two endpoint value hypervectors). To do so, we initially pick a random hypervector to represent the minimal pixel value of 0. Then, starting from the initial value hypervector associated with the minimum pixel value, we generate a new value hypervector for the next pixel value by randomly flipping $\frac{D}{2 \times 255}$ elements of the preceding value hypervector each time. By doing so, we get 256 value hypervectors, including two orthogonal value hypervectors that represent the maximum and minimum pixel values. The 256 value hypervectors are stored in a value memory. The normalized Hamming distance between value hypervectors is shown in Fig. 2(b). It can be observed that the similarity between two value hypervectors is gradually increasing when their associated pixel values decrease. In fact, the normalized Hamming distance between each two value hypervectors is proportional to the corresponding pixel value difference.

Note that once the position hypervectors and value hypervectors are decided and stored in position and value memories, respectively, they will remain fixed and be used throughout the entire training and testing/inference stages [42, 44].

### 3.2 HDC Classifier

Like in conventional classification models [36], a HDC classifier also consists of a training stage and a testing/inference stage, as illustrated in Fig. 3(b).

*3.2.1 Training.* As illustrated in Fig. 3(a), a key step of the HDC classifier is to map a training sample image to a sample hypervector using an encoder based on the image's position/value hypervectors as well as MAP operation. For a sample image in the training dataset, we generate 784 pixel hypervectors, each representing one of the 784 pixels in the image. Specifically, each pixel hypervector is computed by multiplying the corresponding position hypervector and value hypervector. Next, we add up all the 784 pixel hypervectors and binarize the resulting hypervector using the majority rule, thus generating a sample hypervector that represents the sample image in a hyperdimensional space.

The ultimate goal of HDC classifier training is to generate an associative memory that contains 10 class hypervectors associated with 10 labels/classes in the MNIST dataset. To generate a class hypervector, we encode all the sample images in this class into the corresponding sample hypervectors, which are then combined together using the superposition/addition operation. Similarly, the majority rule is adopted to guarantee the class hypervector to be binary. Each class hypervector essentially represents the "center" of all sample hypervectors in that class.

*3.2.2 Testing/Inference.* For testing or inference, using the same encoder as the in the training stage, each new image is first encoded into a query hypervector that represents the image in the considered hyperdimensional space. In conventional machine learning [36], the class whose center is closest to the query in a certain metric space is returned as the prediction label. The same principal applies in the context of HDC classifiers. Specifically, we compare the similarity of the query hypervector to each class hypervector in the associative memory in terms of the (normalized) Hamming distance. Then, the HDC classifier will return the label of the class hypervector, which has the minimum Hamming distance to the query hypervector.

## 4 THREAT MODEL

Our threat model largely follows the common attack scenarios in the context of fooling traditional DNN models [31], while the objective function in Eqn. (2) is tailored to HDC classifiers. Specifically, we consider an adversary (a.k.a. attacker) who minimally modifies a query image in order to make the HDC classifier's prediction erroneous.

**Attacker's capability.** We focus on a grey-box scenario where the attacker is prohibited to access and control the detailed parameters (e.g., position memory, value memory, and associative memory) and the training dataset of the target HDC classifier. Nonetheless, the attacker is allowed to repeatedly send images to the HDC classifier and obtain the corresponding prediction labels. The attacker can add perturbation to the images. In addition, for each image, the attacker is also able to receive the Hamming distances between the image's query hypervector and each class hypervector, which thus forms our grey-box model. Note that only the Hamming distances, not the query or class hypervectors, are revealed to the attacker. In the context of DNN-based image classification, the softmax probability for each class in the last layer as well as the prediction label are typically provided to the user [15]. Thus, our assumption of the attacker's knowing the hamming distances (which are the counterparts of softmax probabilities in DNNs) is not restrictive.

**Attacker's goal.** With limited knowledge about the target HDC classifier, the attacker aims to craft an adversarial image which, with as little perturbation as possible, can mislead the HDC classifier to predict a label other than the true label. This is consistent with non-targeted attacks in standard adversarial machine learning [32, 33].

**Problem formulation.** In the MNIST dataset with $K = 10$ classes, we denote the pixel representation of an input image in a vector form as $X \in \mathbb{R}^{784}$. Then, given the target HDC classifier, we use $\mathbf{f}(X) = [f_1(X), \cdots, f_K(X)] \in [0, 1]^K$ to represent the normalized Hamming distances between the input $X$'s hypervector and the $K$ class hypervectors. Specifically, the HDC classifier first encodes the input $X$ into a hypervector, then computes the Hamming distances $\mathbf{f}(X) = [f_1(X), \cdots, f_K(X)] \in [0, 1]^K$, and finally decides the prediction class label $t_X$ as the one with the minimum Hamming distance.

Given a benign image $X$ with its true class label $t_0$, the attacker would like to create an adversarially perturbed image $\tilde{X} \in \mathbb{R}^{784}$ such that the predicted label $t_{\tilde{X}} = \arg\min_k \{\mathbf{f}(\tilde{X})\}$ for $\tilde{X}$ differs from the true label $t_0$. Formally, we can define the objective function as

$$g(\tilde{X}, t_0) = \max\{\min_{k \neq t_0}[\mathbf{f}(\tilde{X})] - f_{t_0}(\tilde{X}), -\epsilon\}, \quad (1)$$

where $\min_{k \neq t_0}[\mathbf{f}(\tilde{X})]$ is the minimum Hamming distance of the perturbed image to any of the class hypervectors with wrong labels, $f_{t_0}(\tilde{X})$ is the Hamming distance of the perturbed image to the true class hypervector, and a small constant $\epsilon > 0$ is included inside "max{$\cdot$, $-\epsilon$}" to indicate that the attacker does not need to add further perturbation if its attack is already successful (i.e., $\min_{k \neq t_0}[\mathbf{f}(\tilde{X})] - f_{t_0}(\tilde{X})$ is already less than $-\epsilon$). Thus, by minimizing $g(\tilde{X}, t_0)$, the attacker can effectively increase the Hamming distance of the perturbed image to the true class hypervector while decreasing its Hamming distance to other class hypervectors, which hence misleads the HDC classifier to a wrong prediction label.

Meanwhile, the attacker also needs to keep its perturbation to the original image $X$ as minimum as possible, which can be modeled by including regularization terms. Concretely, the attacker obtains $\tilde{X}$ by minimizing the following regularized objective function:

$$\min_{\tilde{X}} \left\{ g(\tilde{X}, t_0) + c \cdot \|\tilde{X} - X\| \right\}, \quad (2)$$

where $\|\tilde{X} - X\|$ is a certain norm that quantifies the difference between $\tilde{X}$ and $X$, and $c \geq 0$ adjusts the weight for regularization. We can also add multiple norms for regularization. For example, $L_0$ norm controls the number of modified pixels, $L_2$ norm controls the squared difference between two images' pixel values, while $L_\infty$ controls the maximum difference between two images' pixel values.

Note that the attacker only knows the Hamming distances $\mathbf{f}(\tilde{X}) = [f_1(\tilde{X}), \cdots, f_K(\tilde{X})]$ and $t_{\tilde{X}} = \arg\min_k\{\mathbf{f}(\tilde{X})\}$, but not the functions $\mathbf{f}(\cdot) = [f_1(\cdot), \cdots, f_K(\cdot)]$ themselves that map an input $\tilde{X}$ to the resulting Hamming distances. As a result, the attacker cannot directly solve the optimization problem in Eqn. (2).

## 5 A MODIFIED GENETIC ALGORITHM

In this section, to create an adversarial input without a white-box HDC classifier, we first describe a basic genetic algorithm and then propose the modifications we make so as to reduce the amount of perturbation introduced to the original benign input.

### 5.1 Genetic Algorithm

Because of the complex Hamming distance function $\mathbf{f}(\cdot)$ and 256 possible values for each of the 784 pixels, the optimization problem in Eqn. (2) involves non-convex integer programming. More importantly, the function $\mathbf{f}(\cdot)$ is non-differentiable and unknown to the attacker. Here, to solve Eqn. (2), we propose a modified genetic algorithm, called Genetic Algorithm with Critical Gene Crossover and Perturbation Adjustment (GA-CGC-PA). Inspired by evolutionary theory and natural selection, genetic algorithm is an iterative search heuristic that has been applied to solve various optimization problems [37, 38, 45].

Concretely, for adversarial attacks, GA-CGC-PA described in Algorithm 1 takes an original input image as an ancestor, from which the first generation of population is generated by natural mutation. A fitness score for each member (i.e., a candidate adversarial image) is evaluated according to a predefined fitness function, which is also the additive inverse of objective function defined in Eqn. (2). Thus, for each member, the attacker needs a query from the HDC classifier to evaluate its fitness score.

Members with higher fitness scores will be selected to breed the subpopulation, forming the next generation. In each generation, natural mutation is also considered. By constantly repeating the evolutionary process, an optimal (possibly locally optimal) individual is ultimately obtained.

A basic genetic algorithm includes four main steps — population initialization, member selection, crossover, and mutation — as described in detail below.

---

**Algorithm 1** Modified Genetic Algorithm (GA-CGC-PA)

**Input:**
Original input $X$, true label $t_0$, population size $N$, maximum iteration $I_{max}$

**Output:** adversarial sample $\tilde{X}$

  1  Create the initial generation $P^0$ from $X$.
  2  $G_{curr} \leftarrow P^0$
  3  **for** $ite = 1$ to $I_{max}$ **do**
  4     Compute fitness score of each member in $G_{curr}$
  5     Find the elite member $Eli$ in $G_{curr}$
        $Eli = \arg\max_{x \in G_{curr}} fitness(x)$
  6     Save $Eli$ as a member of next generation $G_{next}$
  7     **if** $\arg\min_k(f(Eli)) \neq t_0$ **then**
  8         $\tilde{X} \leftarrow Eli$
  9         **return** $\tilde{X}$
 10       **break**
 11     **endif**
 12     Compute selection probability $P_{sel}$ of $G_{curr}$
 13     **for** $num$=2 to N **do**
 14         Choose a pair of parents in $G_{curr}$ according to $P_{sel}$
 15         Apply Critical Gene Crossover (Algorithm 2)
 16         Apply clipping and add clipped child to $G_{next}$
 17     **endfor**
 18     $G_{curr} \leftarrow G_{next}$
 19  **endfor**
 20  Apply Perturbation Adjustment (Algorithm 3)

---

**Algorithm 2** Critical Gene Crossover

**Input:**
$Parent_1$ and $Parent_2$, crossover probability $(p, 1-p)$ of $Parent_1$ and $Parent_2$ with $p > 1-p$, maximum $L_\infty$ mutation distance $\sigma_{max}$, mutation probability $\rho$, critical threshold $\beta$

**Output:** $child$

  1  $child \leftarrow Parent_1$.
  2  Apply $2 \times 2$ max pooling to $child$
       $child' = maxpooling(child)$
  3  Up-sample $child'$ to the original dimension $28 \times 28 = 784$
  4  Normalize values of $child'$ to $[0, 1]$
       $child' = \frac{child' - min(child')}{max(child')}$
  5  Find indexes of critical genes $idx$ such that
       $child'[idx] > \beta$
  6  Update critical genes of $child$
       $child[idx] = p \times Parent_1[idx] + (1-p) \times Parent_2[idx]$
  7  Mutate $child$
       $child[idx] = child[idx] + B(1, \rho) \times \mu(-\sigma_{max}, \sigma_{max})$
  8  **return** $child$

---

*5.1.1 Population Initialization.* The first generation is initialized by applying uniformly distributed random noise in the allowed range $(-\sigma_{max}, \sigma_{max})$ to each gene of the ancestor $X$. For the MNIST dataset, each gene corresponds to one pixel. In total, there are $28 \times 28 = 784$ genes in each individual member, and the algorithm creates $N$ members in each generation.

*5.1.2 Member Selection.* The quality of each population member is evaluated by computing a fitness score according to the fitness function (additive inverse of Eqn. 2). Population members with higher fitness scores are more likely to be selected to reproduce the next generation, whereas members with lower fitness scores are replaced with a higher probability. Towards this end, we compute the softmax of the fitness scores in one generation to obtain the selection probability distribution of the population. We then randomly choose pairs of parents to breed offsprings according to the softmax probability distribution. In order to save the member with the highest fitness score (called *elite* member) in one generation, an elitism technique [46] is employed, where the genes of the elite member are exactly cloned by a member in the next generation.

*5.1.3 Crossover.* Our algorithm makes use of uniform crossover to mate two parents. Each gene of an offspring is produced by combining genes of both parents, $Parent_1$ and $Parent_2$, according to the probability distribution $(p, 1-p)$. We get $p$ through dividing the fitness of the first parent $P_1$ by the sum fitness of both parents. Thus, the child's genes are given as follows:

$$child = p \times Parent_1 + (1-p) \times Parent_2. \tag{3}$$

Nonetheless, since it is required that the perturbation made to the original image be kept as minimum as possible, we reduce the number of perturbed genes (pixels) by using a modified version of uniform crossover, which we call critical gene crossover as described in Section 5.2.1.

*5.1.4 Mutation.* In order to promote diversity within a generation and improve the search power of the genetic algorithm, the child generated by crossover has to be mutated and clipped before becoming a member of the next generation. Like population initialization, random noise is sampled uniformly from a range $(-\sigma_{max}, \sigma_{max})$ and added to the chromosome of the child with a mutation probability $\rho$. Considering that a feasible solution has to possess a reasonable gene (e.g. pixel value for MNIST dataset), a mutated child is clipped to ensure that its genes are all within an allowable range.

## 5.2 Modification for Perturbation Reduction

While the basic genetic algorithm can generate an adversarial image to fool the HDC classifier, the amount of perturbation can be really significant (see Fig. 4(b) for an example), making the adversarial input more easily identified by human perception. Here, we propose to use *critical gene crossover* and *perturbation adjustment* to significantly reduce the amount of perturbation.

*5.2.1 Critical Gene Crossover.* While each individual member carries a large number of genes (784 genes each MNIST image), not all the genes are equally critical to the improvement of fitness score. Additionally, the standard uniform crossover modifies each pixel of the original image, which unnecessarily introduces redundant perturbation. To reduce perturbation, we propose critical gene crossover to selectively cross the parents' most important genes. To do so, we first make a child by duplicating the parent with the higher fitness score and then select critical genes using the max pooling operation. Next, we renew the critical genes by uniformly

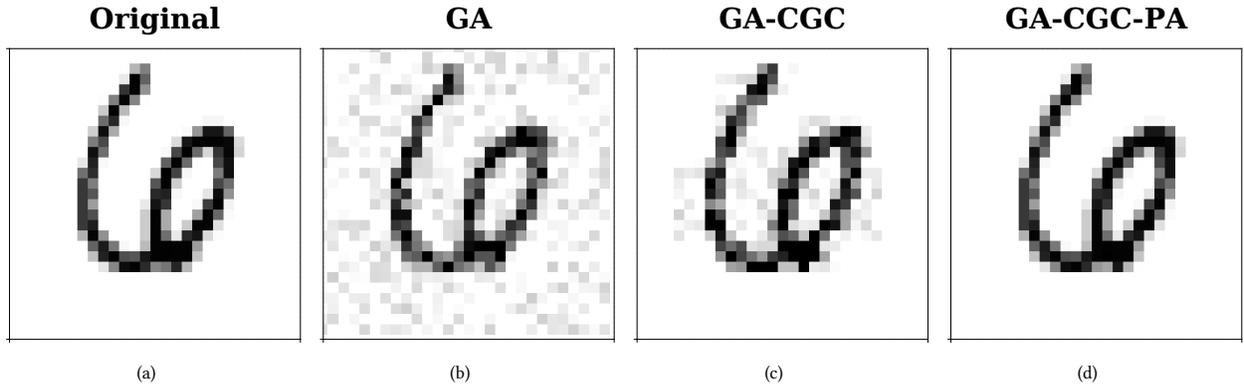

**Figure 4: Comparison of different adversarial attacks that mislead the HDC classifier to classify "6" as "2". (a) Original benign image. (b) Adversarial image by basic genetic algorithm (GA). (c) Adversarial image by genetic algorithm with critical gene crossover (GA-CGC). (d) Adversarial image by our proposed genetic algorithm with critical gene crossover and perturbation adjustment (GA-CGC-PA).**

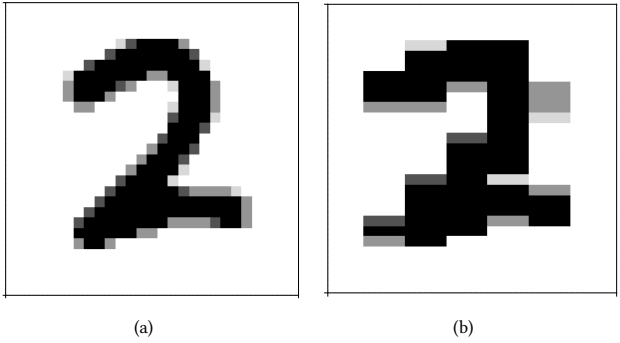

**Figure 5: Importance sampling to select critical genes. (a) The original image of digit "2", with perturbation possibly added on $28 \times 28 = 784$ pixels (genes). (b) After importance sampling via $2 \times 2$ max pooling, perturbation is restricted to critical genes (i.e., pixels) with sufficiently large pixel values.**

crossing those of the two parents. The detailed steps are described in Algorithm 2.

We define critical genes as the ones that mostly differentiate images of different classes. For the example of the MNIST dataset, pixels that are close to and form the digit are more important than others that have lower pixel values and mostly form the background, and hence can be chosen as critical genes. This can be done via importance sampling [47] and is illustrated in Fig. 5. Concretely, we run a $2 \times 2$ max pooling of the absolute pixel value for each image, up-sample to the original dimension and normalize all the values to the range of $[0, 1]$. Then, we execute Boolean comparison of all the values to a threshold $\beta$ and decide the critical genes.

*5.2.2 Perturbation Adjustment.* Considering the fact that the genetic algorithm generates random mutation in each generation and thus can introduce unnecessary modification to the original image, we propose to further reduce the perturbation by using perturbation adjustment while still keeping the adversarial attack

---

**Algorithm 3** Perturbation Adjustment

**Input:**
Original image $X$, true label $t_0$, adversarial image $\tilde{X}$

1 Find an index list $\mathcal{L}$ for pixels that differ in $X$ and $\tilde{X}$
2 **for** $p$ in $\mathcal{L}$ **do**
3      $v_{ori} \leftarrow X[p]$
4      $v_{adv} \leftarrow \tilde{X}[p]$
5      **for** $v = v_{ori}$ to $v_{adv}$ **do**
6          $\tilde{X}[p] = v$
7          **if** $\arg\min_k (\mathbf{f}(\tilde{X})) \neq t_0$ **then**
8              **break**
9          **endif**
10      **endfor**
11 **endfor**

---

successful. Our perturbation adjustment technique is described in Algorithm 3. It starts by finding an index list $\mathcal{L}$ of modified pixels in the adversarial image compared to the original image. For each pixel in the list $\mathcal{L}$, its value is restored to the original value $v_{ori}$. Then, we gradually change the value towards the adversarial value $v_{adv}$ in the adversarial image and stop this process until the adversarial image can successfully mislead the HDC classifier to a wrong prediction.

## 5.3 Effect of Perturbation Reduction

To highlight the effect of our proposed modification — critical gene crossover and perturbation adjustment — to the basic genetic algorithm, we present an example of adversarial attacks on the digit "6" using three different algorithms in Fig. 4: standard genetic algorithm without modification (GA), modified genetic algorithm with only critical gene crossover (GA-CGC), and modified genetic algorithm with both critical gene crossover and perturbation adjustment (GA-CGC-PA). The HDC classifier is trained on the MNIST dataset as described in Section 6.1. In all the three attacks, the HDC classifier misclassifies the digit "6" as "2". Fig. 4(a) shows the original benign image for digit "6" which can be correctly classified by the HDC classifier, while Fig. 4(b) shows the adversarial image using GA. We can clearly see that many pixels in the original image are modified and added with perturbation noise, making the adversarial image easily identifiable. Fig. 4(c) shows the adversarial image generated by GA-CGC after using critical gene crossover. Compared with the result in Fig. 4(b), many background pixels in Fig. 4(c) are left unchanged and only pixels surrounding the digit are altered. By using GA-CGC-PA with further perturbation adjustment, the adversarial image is shown in Fig. 4(d), which looks very similar to the original benign image but is still misclassified by the HDC classifier as "2". This shows the clear advantage of GA-CGC-PA over the basic genetic algorithm and only using critical gene crossover, in terms of reducing the amount of perturbation in adversarial images.

**Table 1: Comparison of Three Adversarial Attacks**

| Attack | GA | GA-CGC | GA-CGF-PA |
|---|---|---|---|
| # Modified Pixels | 438 | 218 | 9 |
| $L_2$-distance | 3.491 | 2.245 | 0.187 |
| $L_\infty$-distance | 0.298 | 0.296 | 0.160 |

Further, we provide in Table 1 a quantitative comparison of perturbation introduced by the three algorithms in terms of the $L_p$ norm metric. From the result, we see that the number of pixels modified (i.e., $L_0$ norm) is largely reduced from 438 to 9 by using GA-CGC-PA. In addition, with GA-CGC-PA, the $L_2$ and $L_\infty$ norms of the perturbation also decrease significantly compared to the basic GA and GA-CGC.

## 6 EVALUATION RESULTS

This section validates the effectiveness of our proposed GA-CGC-PA for adversarial attacks on a target HDC classifier. We focus on the MNIST dataset and train a HDC classifier based on the design in Section 3. Then, we show that: with a random majority rule, GA-CGC-PA can significantly reduce the classification accuracy for all test images; and with a fixed majority rule, GA-CGC-PA can successfully make the HDC classifier misclassify 78% of the otherwise correctly-classified benign images. Importantly, our results are the first to highlight that the emerging HDC classification models are vulnerable to adversarial attacks.

### 6.1 HDC Classifier Training

We train a HDC classifier based on the MNIST training dataset [35], while noting that it is an active research topic in the field to design HDC classifiers for more sophisticated datasets like ImageNet [11, 19]. We create 784 position hypervectors to represent $28 \times 28 = 784$ pixel positions and 256 value hypervectors to represent pixel values. The dimensionality for each hypervector is $D = 10^4$. Then, as described in Section 3, we encode each training sample into a sample hypervector and obtain 10 class hypervectors based on the training dataset. Next, we project each test image into a query hypervector and compare it against class hypervectors. Recalling that in the hypervector encoding process, we use the majority rule for vector binarization. By using the random majority rule (RMR) that randomly assigns 1 or −1 in the rare event that the sum is zero after superposition operation, the HDC classifier may assign different labels in different inferences for the same input.

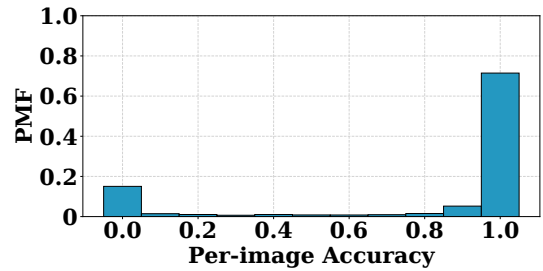

**Figure 6: Probability mass function (PMF) of per-image accuracy on the HDC classifier with RMR.**

To eliminate this uncertainty, we can also apply the fixed majority rule (FMR) that always assigns 1 or −1.

For the HDC classifier with RMR, we execute 1,000 rounds of classification for each test image to calculate the average accuracy, which we also refer to as *per-image* accuracy. Fig. 6 shows the distribution of per-image accuracy for the test dataset. We see that, for around 70% of the test images, the HDC classifier can assign correct labels with 100% per-image accuracy. Meanwhile, there are about 15% test images that are almost always misclassified. For the remaining 15% test images, the HDC classifier behaves unconfidently and sometimes yields misclassified results. Consequently, the test images that have 100% per-image accuracy are harder to attack (called *hard* cases) than those with a lower per-image accuracy (called *vulnerable* case). In other words, vulnerable images can be considered already "adversarial" to our HDC classifier to some extent, although they are from the MNIST dataset. The overall accuracy of our HDC classifier is lower than that of DNNs [15], and can be improved by enlarging the MNIST dataset, which is beyond the scope of our work. Importantly, as we will show later, GA-CGC-PA can successfully mislead the HDC classifier with a high probability regardless of hard or vulnerable cases.

### 6.2 Attack on HDC Classifier with RMR

We first evaluate GA-CGC-PA by considering the random majority rule (RMR) for the HDC classifier. In our algorithm, we use population size $N = 6$, mutation probability $\rho = 0.05$, max pooling size 2×2, and critical threshold $\beta = 0$.

We focus on attacking the hard cases (i.e., those images with 100% per-image accuracy), while noting that the already-vulnerable images (i.e., those with less than 100% per-image accuracy) are even easier to attack. Fig. 7 visually illustrates the benign input images, adversarial perturbation noise, and the corresponding adversarial images. The adversarial images can significantly decrease the HDC classifier's performance, while they are still barely recognizable by human eyes. Moreover, because of the critical gene crossover, adversarial perturbation noises are mostly added around the digit pixels rather than spread throughout the whole image.

Fig. 8 shows the evolution of fitness scores for attacking the benign images of digits "0" and "1" (in Fig. 7) when using GA-CGC-PA. Each iteration produces a new population generation (i.e., a set of 6 candidate adversarial images). We can see that GA-CGC-PA gradually increases the fitness score (which is the additive

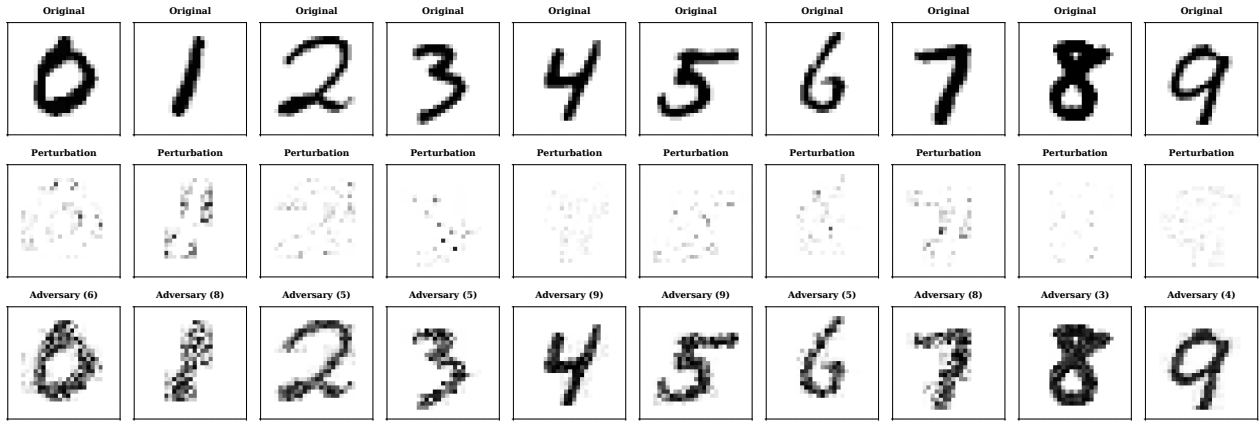

**Figure 7: Adversarial attacks on the HDC classifier with RMR. The first row shows the original benign images. The second row shows the perturbation noise added by the attacker. The third row shows the adversarial images, and the corresponding misclassified labels are given at the top of each image.**

**Table 2: Perturbation for Images Shown in Fig. 7**

| Digit | 0 | 1 | 2 | 3 | 4 | 5 | 6 | 7 | 8 | 9 |
|---|---|---|---|---|---|---|---|---|---|---|
| # Modified Pixels | 289 | 160 | 260 | 196 | 185 | 241 | 177 | 192 | 256 | 253 |
| $L_2$-distance | 4.626 | 5.073 | 3.02 | 3.703 | 1.37 | 3.028 | 2.58 | 4.017 | 2.029 | 1.537 |
| $L_\infty$-distance | 0.867 | 0.968 | 0.643 | 0.956 | 0.276 | 0.653 | 0.737 | 0.92 | 0.401 | 0.271 |

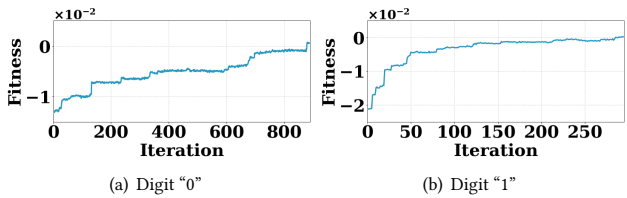

**Figure 8: Fitness score for digits "0" and "1" in Fig. 7. Given a population size $N = 6$ in each generation in GA-CGC-PA, the attacker needs 6 queries in each iteration.**

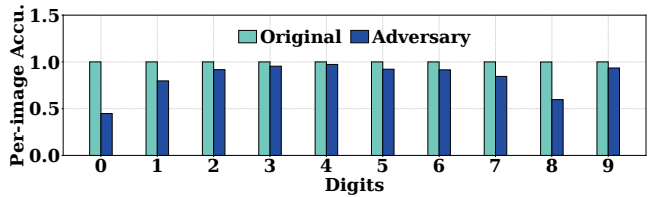

**Figure 9: Per-image accuracies of benign images and corresponding adversarial images shown in Fig. 7.**

inverse of our objective function in Eqn. 2), thus iteratively updating adversarial images.

Next, we show the corresponding per-image accuracies of both original images and adversarial ones in Fig. 9. It can be clearly seen that, with GA-CGC-PA, all the images become vulnerable with a per-accuracy lower than 100%. In particular, the sample images for digits "0" and "8" in Fig. 7 have the lowest accuracy after attacks and hence are relatively easier to attack than others.

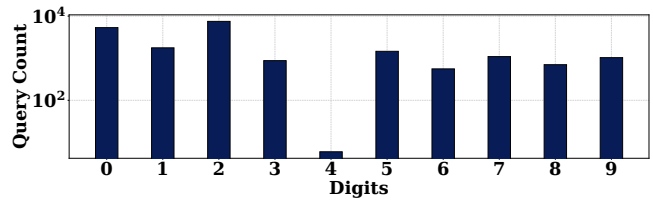

**Figure 10: Query counts needed to generate adversarial images shown in Fig. 7.**

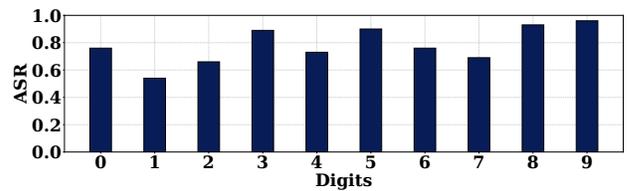

**Figure 11: ASR of digits 0-9 for HDC classifier with FMR.**

### 6.2.1 Amount of Perturbation.
Next, we quantify the adversarial perturbation noise generated and added to the benign images. To have a successful attack, the adversarial images need to not only deceive the HDC classifier but also have as small perturbation as possible compared to benign ones. To this end, the amount of perturbation is an important metric to evaluate the attack algorithm. As in the prior studies on adversarial machine learning [33], we use $L_0$ norm, $L_2$ norm, and $L_\infty$ norms to measure the amount of

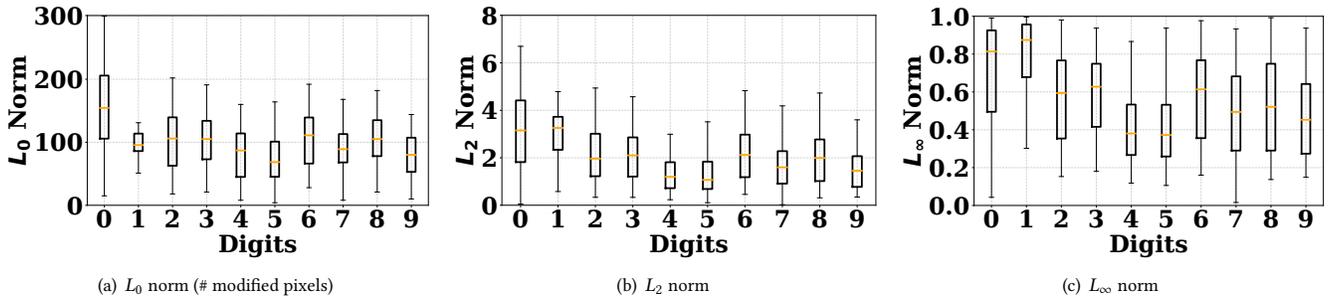

(a) $L_0$ norm (# modified pixels)

(b) $L_2$ norm

(c) $L_\infty$ norm

**Figure 12: Box plot of perturbation noise added by** GA-CGC-PA **for the HDC classifier with FMR. Each box plot shows the values for the maximum/minimum/median/75th percentile/25th percentile, excluding outliers.**

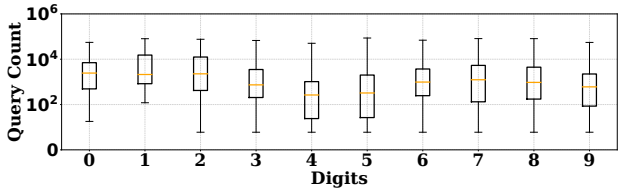

**Figure 13: Box plot of query count needed by** GA-CGC-PA **for the HDC classifier with FMR. Each box plot shows the values for the maximum/minimum/median/75th percentile/25th percentile, excluding outliers.**

perturbation. Note that while $L_0$ is not a mathematical norm, it is commonly used to quantify the total number of modified pixels in our context. By definition, $L_2$ norm indicates the overall perturbation noise added to a benign image, while $L_\infty$ norm measures the maximum per-pixel perturbation noise.

Table 2 shows the three norm distances for the perturbation noise added to the benign images shown in Fig. 7. It is worth noting that $L_2$ and $L_\infty$ norms are calculated over the images with normalized pixel values in the range of $[0, 1]$. For all the adversarial images shown in Fig. 7, there are fewer than 300 modified pixels. While the $L_\infty$ is large, the $L_2$ norm is reasonably small for most digits, indicating the overall perturbation added by GA-CGC-PA is not large, which can also be observed from Fig. 7.

*6.2.2 Query Count.* In addition, query efficiency is also an important metric to evaluate a black-box or grey-box attack, since access to the target classifier may be limited. From the attacker's perspective, fewer queries are desired in order to hide its identity and stealthiness for attacks. Thus, we plot in Fig. 10 the number of queries used to generate the adversarial images.

The result shows that the average query count is up to the order of thousands. In particular, the query count for digit "2" is more than 7k, whereas the digit "4" needs the least number of queries to attack. While the existing adversarial attacks in the literature focus on DNN-based classifiers and different datasets, we note that they typically need an order of 10k or more queries to successfully attack an image [37, 38].

## 6.3 Attack on HDC Classifier with FMR

We now turn to the fixed majority rule (FMR) such that the prediction label for a given image is fixed without uncertainties. The hyperparameters for GA-CGC-PA are the same as in Section 6.2.

*6.3.1 Attack Success Rate.* With FMR, the per-image accuracy is either 0 or 1. Thus, instead of considering an individual image, we demonstrate the attack success rate (ASR) of GA-CGC-PA over multiple images. Specifically, we randomly pick 200 correctly classified images for each digit from "0" to "9" from the MNIST dataset. For each image, we apply GA-CGC-PA to generate the corresponding adversarial image subject to a maximum query count of $10^5$ (i.e., $I_{max} = 10^5$ in Algorithm 1). If an adversarial image is successfully generated to fool the HDC classifier within the query limit, it is regarded as a successful attack, and a failed attack otherwise.

We compute the ASR over 200 images for each digit and present the results in Fig. 11. It can be seen that GA-CGC-PA is successful for all the digits in most cases, with digits "3", "5", "8" and "9" having the highest ASR. Considering the 10 digits altogether, we obtain an average ASR of 0.78.

*6.3.2 Amount of Perturbation.* To show the amount of adversarial perturbation added by GA-CGC-PA, we provide the bar plot of perturbation amount in terms of $L_0$, $L_2$ and $L_\infty$ norms in Fig. 12. As one can see from the figure, the median number of modified pixels for most adversarial images is around 100. The $L_2$ norm for the majority of perturbation noise is between 2 and 4, whereas the $L_\infty$ norm lies mostly between 0.3 and 0.8 for most images.

*6.3.3 Query Count.* Next, we calculate the query counts for the successfully attacked images and show the results in a box plot in Fig. 13. We can notice that the median query count of all digits is less than 5,000, which is a reasonably good query efficiency for black-/grey-box attacks [37].

*6.3.4 Adversarial Examples.* Finally, we visually show some adversarial examples for the HDC classifier with FMR, and provide the corresponding amount of perturbation noises. We choose two sets of examples: hard case and vulnerable case. In the hard case, benign images would have a 100% per-image accuracy had the HDC classifier use RMR. In the vulnerable case, benign images are correctly classified by the HDC classifier with FMR, but would have less than 100% per-image accuracy had the classifier use RMR. That is, the vulnerable images are those borderline images that are already hard to correctly classify by the HDC classifier.

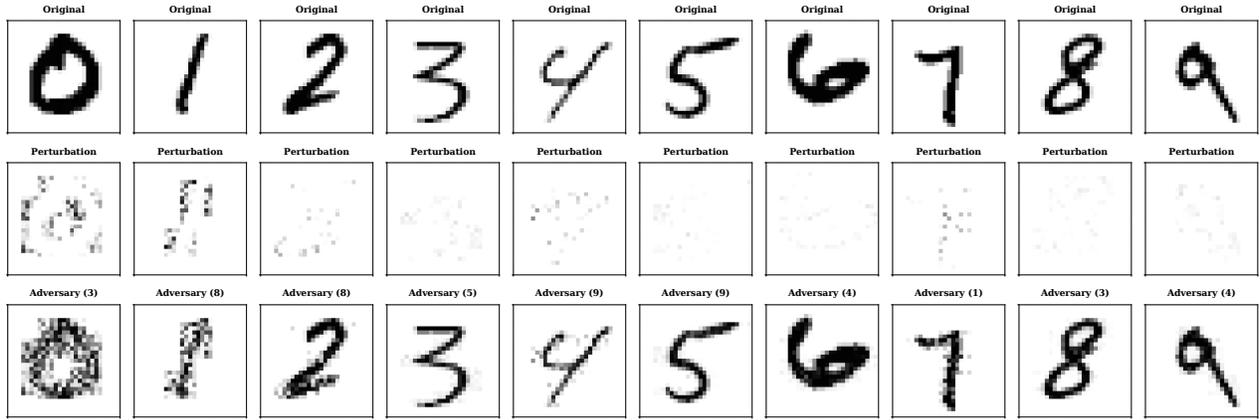

**Figure 14: Adversarial attacks on the HDC classifier with FMR (hard case). The first row shows the original benign images. The second row shows the perturbation noise added by the attacker. The third row shows the adversarial images, and the corresponding misclassified labels are given at the top of each image.**

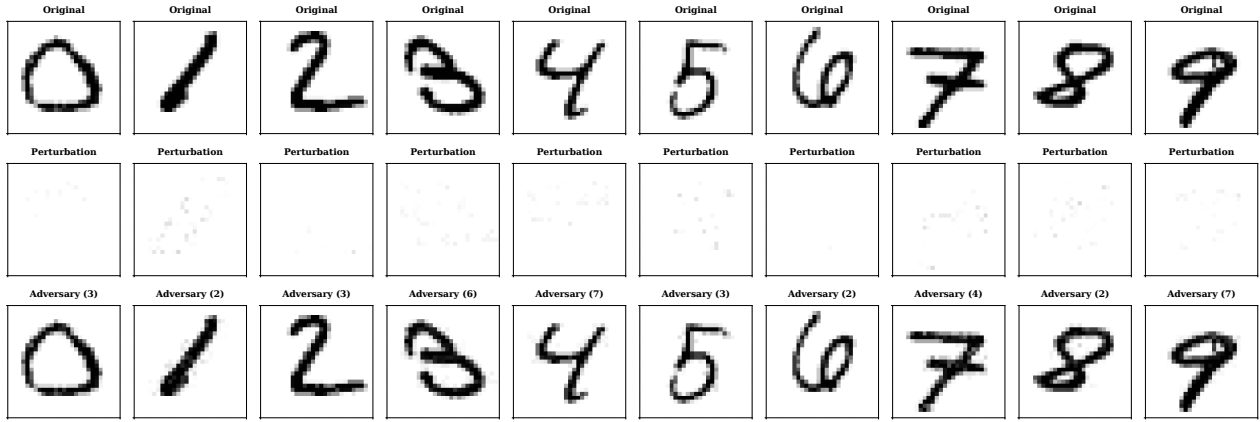

**Figure 15: Adversarial attacks on the HDC classifier with FMR (vulnerable case). The first row shows the original benign images. The second row shows the perturbation noise added by the attacker. The third row shows the adversarial images, and the corresponding misclassified labels are given at the top of each image.**

**Table 3: Perturbation for Images Shown in Fig. 14 and Fig. 15. The values for Fig. 15 are shown in parentheses.**

| Digit | 0 | 1 | 2 | 3 | 4 | 5 | 6 | 7 | 8 | 9 |
|---|---|---|---|---|---|---|---|---|---|---|
| # Modified Pixels | 301 (22) | 123 (82) | 95 (21) | 104 (107) | 72 (42) | 117 (34) | 113 (9) | 74 (86) | 114 (68) | 97 (45) |
| $L_2$-distance | 6.73 (0.36) | 4.48 (0.95) | 2.20 (0.45) | 0.82 (0.65) | 1.54 (0.44) | 0.69 (0.66) | 0.76 (0.18) | 1.90 (1.12) | 0.64 (0.73) | 0.72 (0.49) |
| $L_\infty$-distance | 0.92 (0.21) | 0.96 (0.24) | 0.69 (0.21) | 0.21 (0.15) | 0.53 (0.13) | 0.19 (0.21) | 0.23 (0.16) | 0.56 (0.37) | 0.18 (0.25) | 0.20 (0.19) |

The benign images, perturbation noise, and adversarial images for hard and vulnerable cases are shown in Fig. 14 and Fig. 15, respectively. Also, we give the amount of perturbation noises for the two cases in Table 3.

It is more difficult to launch successful attacks in the hard case than in the vulnerable case. Thus, as expected, the perturbation noise added by GA-CGC-PA in the hard case is generally less than in the vulnerable case. In particular, in the vulnerable case, the adversarial image is almost identical to the corresponding benign image by human perception. This can also be reflected from the perturbation noise figures and Table 3.

## 7 DEFENSE STRATEGIES

We highlight two simple defense strategies — adversarial training (also commonly used for defending DNNs) and retraining (relevant to ensemble learning) — to degrade the attack success rate (ASR).

## 7.1 Adversarial Training

Adversarial training is a common and proven defense approach to increasing the robustness of classification models, especially for the recently exploding DNN-based classifiers [33, 48, 49]. The key idea is to strengthen the classifier by augmenting training data with adversarial samples throughout the training process. Similarly, this strategy can be also applied in defending HDC classifiers.

To evaluate the effect of adversarial training in defending HDC classifiers, we first generate 2,000 adversarial samples, 200 for each digit, based on the HDC classifier with FMR. Combining the adversarial samples with the original training dataset, we retrain the HDC classifier with the same set of position hypervectors and value hypervectors as before, and obtain the corresponding associative memory containing 10 class hypervectors. Next, we use the 2,000 adversarial samples to attack the new HDC classifier and get the ASR shown in Fig. 16. Note that unlike the ASR in Section 6.3, we focus on whether the new HDC classifier still misclassify the adversarial samples that are misclassified by the original classifier. Thus, we compute the ASR by dividing the number of still misclassified samples by the total number of adversarial 2,000 samples misclassified by the original HDC classifier. As a result, the ASR for the original HDC classifier is considered as 1. We can see that ASR to the HDC classifier with adversarial training decreases significantly to around 0.5, which means that only half of the adversarial samples can still mislead the HDC classifier. While the attacker can further modify adversarial samples to possibly fool the new HDC classifier again, adversarial training clearly are effective against some adversarial images and hence increases the difficulty of successful attacks.

A potential drawback of adversarial training is that many adversarial samples need to be generated and included into the training dataset, which might degrade the classification accuracy for benign images as shown in DNN-based classifiers [15, 31, 37].

## 7.2 Retraining

Random indexing (Section 2.1) is a unique feature in HDC classifiers. By random indexing, different position hypervectors and value hypervectors are generated, resulting in different class hypervectors (i.e., different HDC classifiers) even with the same training dataset. Note that the randomly generated position hypervectors and value hypervectors are fixed for a given HDC classifier. This is similar to ensemble learning [15, 36] where multiple models are trained by using different hyperparameters, but random indexing can create a lot more randomness due to its hyperdimensional space [13] than varying the hyperparameters in traditional machine learning.

The adversarial samples crated by GA-CGC-PA are targeted at a fixed (but unknown to the attacker) set of position hypervectors, memory hypervectors and the trained class hypervectors. Due to the large randomness in these hypervectors, adversarial samples are not expected to transfer well between different HDC classifiers even though they are trained on the same training dataset. We still use the 2,000 adversarial samples for evaluation. Using the same training dataset but a new random indexing, we can train a new HDC classifier. As illustrated in Fig. 16, the ASR for these 2,000 adversarial samples now decreases to less than 0.1, which clearly increases the difficulty to fool the HDC classifier.

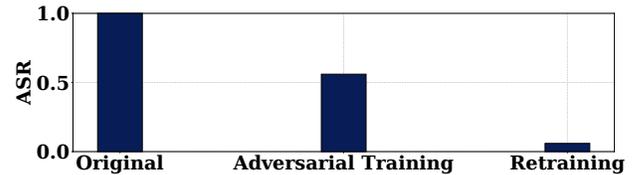

**Figure 16: ASR for the original HDC classifier, and new HDC classifiers with adversarial training and retraining defenses.**

Moreover, based on the idea of moving target defense [50], the defender can train multiple HDC classifiers (by using multiple different position hypervectors and value hypervectors) and randomize the choice of classifiers for each query image. This can further strengthen the security of HDC classifiers, although the storage complexity for inference also increases.

Given constraints such as inference accuracy and complexity, the defender can select an appropriate defense strategy for its HDC classifier. While there can be more advanced defense mechanisms (e.g., detecting adversarial inputs [51]) which we leave as a future research direction, our study is important in that it is the first to demonstrate the vulnerability of emerging HDC classifiers and highlight the urgency of security awareness.

## 8 RELATED WORKS

Adversarial machine learning has been extensively studied in recent years [30, 31, 33, 52, 53]. Adversarial samples can be fed to a target machine learning model during either training or testing/inference stages to degrade the model performance [33, 54]. Here, we consider creating adversarial samples during the inference stage. While early studies focus on generating adversarial images, recent adversarial machine learning has also been extended to adversarial audio signals [29], adversarial texts [55], fooling LiDAR sensors in autonomous driving [56], among others.

In general, adversarial attacks on DNNs can be categorized into white-box attacks, black-box attacks, and grey-box attacks [33]. In a white-box attack, an attacker is assumed to know complete details about the target DNNs [52, 53], which is often too strong in practice. By contrast, in a black-box attack, only benign inputs and the corresponding prediction label (plus some additional information such as softmax probabilities in a grey-box setting) are available to the attacker [38, 57]. For black-box or grey-box attacks, one approach is to obtain a substitute model for the target DNN and then generate adversarial samples offline [57, 58]. Nonetheless, this approach is that it often requires a prohibitively large number of queries to train a substitute DNN model [59–61].

More recent studies on black-box or grey-box attacks have proposed to use gradient estimations to generate adversarial samples [47, 62]. Nonetheless, these approaches are generally limited to differentiable objective functions, which is not the case in HDC classifiers that use Multiply-Add-Permute operation in a hyperdimensional space without differentiable objective functions.

Boundary attack is a gradient-free black-box attack, which creates adversarial samples using an already-available adversarial sample as a reference [63, 64]. Nonetheless, an adversarial sample is needed at the first place. Genetic algorithm is another effective approach to attacks on DNNs [37, 38, 65]. We leverage a genetic algorithm, but also extend it with critical gene crossover and perturbation adjustment to reduce perturbation (see Fig. 4).

Most importantly, we focus on the emerging HDC classifiers that operate on hypervectors and have recently shown promise in increasingly more applications, including image classification [11, 24], seizure onset detection [27], fault isolation in industrial systems [26], and robot navigation [28].

The existing studies on HDC classifiers have been predominantly focused on improving the energy efficiency, inference latency, privacy preservation, or architecture design [14, 16–21, 34, 66]. Nonetheless, adversarial attacks on HDC classifiers have been neglected, raising serious concerns with their safety as they are being adopted in increasing more applications including mission-critical scenarios [11, 26–28]. Our study bridges the gap and demonstrates that, like their DNN counterparts, HDC classifiers can be vulnerable to adversarial inputs and hence need to be better safeguarded.

## 9 CONCLUSION

In this paper, we study adversarial attacks on emerging HDC classifiers. We first build a HDC classifier for image classification on the MNIST dataset, and formulate a regularized optimization problem for grey-box attacks, where the attacker's goal is to generate adversarial images for misclassification. Then, we propose a modified genetic algorithm (GA-CGC-PA) to generate adversarial images within a reasonably small number of queries. Our results show that GA-CGC-PA can successfully mislead the HDC classifier to wrong prediction labels with a large probability. Finally, we present two defense strategies — adversarial training and retraining — to safeguard HDC classifiers.

## REFERENCES

[1] W. Shi, J. Cao, Q. Zhang, Y. Li, and L. Xu, "Edge computing: Vision and challenges," *IEEE Internet of Things Journal*, vol. 3, pp. 637–646, Oct 2016.

[2] W. Jiang, X. Zhang, E. H.-M. Sha, L. Yang, Q. Zhuge, Y. Shi, and J. Hu, "Accuracy vs. efficiency: Achieving both through fpga-implementation aware neural architecture search," in *DAC*, 2019.

[3] H. Cai, C. Gan, and S. Han, "Once for all: Train one network and specialize it for efficient deployment," in *ICLR*, 2019.

[4] A. Ren, T. Zhang, S. Ye, J. Li, W. Xu, X. Qian, X. Lin, and Y. Wang, "Admm-nn: An algorithm-hardware co-design framework of dnns using alternating direction methods of multipliers," in *ASPLOS*, 2019.

[5] S. S. Ogden and T. Guo, "Characterizing the deep neural networks inference performance of mobile applications," in *arXiv*, 2019, https://arxiv.org/abs/1909.04783.

[6] W. Liu, X. Ma, S. Lin, S. Wang, X. Qian, X. Lin, Y. Wang, and B. Ren, "Patdnn: Achieving real-time DNN execution on mobile devices with pattern-based weight pruning," in *ASPLOS*, 2020.

[7] X. Ma, F.-H. Guo, W. Niu, X. Lin, J. Tang, K. Ma, B. Ren, and Y. Wang, "Pconv: The missing but desirable sparsity in DNN weight pruning for real-time execution on mobile device," in *AAAI*, 2020.

[8] S. Han, H. Mao, and W. J. Dally, "Deep compression: Compressing deep neural networks with pruning, trained quantization and huffman coding," in *ICLR*, 2016.

[9] Q. Lu, W. Jiang, X. Su, Y. Shi, and J. Hu, "On neural architecture search for resource-constrained hardware platforms," in *ICCAD*, 2019.

[10] T. Elsken, J. H. Metzen, and F. Hutter, "Neural architecture search: A survey.," *Journal of Machine Learning Research*, vol. 20, no. 55, pp. 1–21, 2019.

[11] L. Ge and K. K. Parhi, "Classification using hyperdimensional computing: A review," *IEEE Circuits and Systems Magazine*, vol. 20, no. 2, pp. 30–47, 2020.

[12] G. Karunaratne, M. Le Gallo, G. Cherubini, L. Benini, A. Rahimi, and A. Sebastian, "In-memory Hyperdimensional Computing," *Nature Electronics*, Jun 2020.

[13] P. Kanerva, "Hyperdimensional computing: An introduction to computing in distributed representation," *Cognitive Computation*, vol. 1, pp. 139–159, 2009.

[14] S. Salamat, M. Imani, B. Khaleghi, and T. Rosing, "F5-hd: Fast flexible fpga-based framework for refreshing hyperdimensional computing," in *FPGA*, 2019.

[15] I. Goodfellow, Y. Bengio, and A. Courville, *Deep Learning*. MIT Press, 2016. http://www.deeplearningbook.org.

[16] M. Imani, J. Morris, J. Messerly, H. Shu, Y. Deng, and T. Rosing, "BRIC: Locality-based encoding for energy-efficient brain-inspired hyperdimensional computing," in *DAC*, 2019.

[17] M. Imani, J. Messerly, F. Wu, W. Pi, and T. Rosing, "A binary learning framework for hyperdimensional computing," in *DATE*, 2019.

[18] M. Imani, A. Rahimi, D. Kong, T. Rosing, and J. M. Rabaey, "Exploring hyperdimensional associative memory," in *HPCA*, 2017.

[19] M. Imani, S. Salamat, S. Gupta, J. Huang, and T. Rosing, "Fach: Fpga-based acceleration of hyperdimensional computing by reducing computational complexity," in *ASPDAC*, 2019.

[20] M. Imani, C. Huang, D. Kong, and T. Rosing, "Hierarchical hyperdimensional computing for energy efficient classification," in *DAC*, 2018.

[21] S. Benatti, F. Montagna, V. Kartsch, A. Rahimi, D. Rossi, and L. Benini, "Online learning and classification of emg-based gestures on a parallel ultra-low power platform using hyperdimensional computing," *IEEE Transactions on Biomedical Circuits and Systems*, vol. 13, no. 3, pp. 516–528, 2019.

[22] A. Rahimi, P. Kanerva, and J. M. Rabaey, "A robust and energy-efficient classifier using brain-inspired hyperdimensional computing," in *ISLPED*, 2016.

[23] M. Imani, J. Hwang, T. Rosing, A. Rahimi, and J. M. Rabaey, "Low-power sparse hyperdimensional encoder for language recognition," *IEEE Design Test*, vol. 34, no. 6, pp. 94–101, 2017.

[24] C.-Y. Chang, Y.-C. Chuang, and A.-Y. A. Wu, "Task-projected hyperdimensional computing for multi-task learning," in *Artificial Intelligence Applications and Innovations*, 2020.

[25] E. Chang, A. Rahimi, L. Benini, and A. A. Wu, "Hyperdimensional computing-based multimodality emotion recognition with physiological signals," in *IEEE International Conference on Artificial Intelligence Circuits and Systems*, 2019.

[26] D. Kleyko, E. Osipov, N. Papakonstantinou, and V. Vyatkin, "Hyperdimensional computing in industrial systems: The use-case of distributed fault isolation in a power plant," *IEEE Access*, vol. 6, pp. 30766–30777, May 2018.

[27] A. Burrello, K. Schindler, L. Benini, and A. Rahimi, "Hyperdimensional computing with local binary patterns: One-shot learning of seizure onset and identification of ictogenic brain regions using short-time ieeg recordings," *IEEE Transactions on Biomedical Engineering*, vol. 67, no. 2, pp. 601–613, 2020.

[28] A. Mitrokhin, P. Sutor, C. Fermüller, and Y. Aloimonos, "Learning sensorimotor control with neuromorphic sensors: Toward hyperdimensional active perception," *Science Robotics*, vol. 4, no. 30, 2019.

[29] N. Carlini and D. Wagner, "Audio adversarial examples: Targeted attacks on speech-to-text," in *2018 IEEE Security and Privacy Workshops*, 2018.

[30] X. Yuan, P. He, Q. Zhu, and X. Li, "Adversarial examples: Attacks and defenses for deep learning," *IEEE Transactions on Neural Networks and Learning Systems*, vol. 30, no. 9, pp. 2805–2824, 2019.

[31] L. Huang, A. D. Joseph, B. Nelson, B. I. Rubinstein, and J. D. Tygar, "Adversarial machine learning," in *ACM Workshop on Security and Artificial Intelligence*, 2011.

[32] C. Guo, J. R. Gardner, Y. You, A. G. Wilson, and K. Q. Weinberger, "Simple black-box adversarial attacks," in *ICML*, 2019.

[33] K. Ren, T. Zheng, Z. Qin, and X. Liu, "Adversarial attacks and defenses in deep learning," *Elsevier Engineering*, vol. 6, no. 3, pp. 346 – 360, 2020.

[34] M. Imani, S. Bosch, M. Javaheripi, B. Rouhani, X. Wu, F. Koushanfar, and T. Rosing, "SemiHD: Semi-supervised learning using hyperdimensional computing," in *ICCAD*, 2019.

[35] Y. LeCun, C. Cortes, and C. J. C. Burges, "The MNIST database of handwritten digits," http://yann.lecun.com/exdb/mnist/.

[36] M. Mohri, A. Rostamizadeh, and A. Talwalkar, *Foundations of Machine Learning*. MIT Press, 2018.

[37] M. Alzantot, Y. Sharma, S. Chakraborty, H. Zhang, C.-J. Hsieh, and M. B. Srivastava, "GenAttack: Practical black-box attacks with gradient-free optimization," in *Genetic and Evolutionary Computation Conference*, 2019.

[38] X. Liu, Y. Luo, X. Zhang, and Q. Zhu, "A Black-box Attack on Neural Networks Based on Swarm Evolutionary Algorithm," *Elsevier Computers & Security*, vol. 85, pp. 89–106, August 2019.

[39] S. Datta, "A binary, 2048-dim. generic hyper-dimensional processor," Master's thesis, EECS Department, University of California, Berkeley, May 2019.

[40] T. A. Plate, "Holographic reduced representations," *IEEE Transactions on Neural networks*, vol. 6, no. 3, pp. 623–641, 1995.

[41] E. P. Frady, D. Kleyko, and F. T. Sommer, "A theory of sequence indexing and working memory in recurrent neural networks," *Neural computation*, vol. 30, no. 6, pp. 1449–1513, 2018.

[42] D. Kleyko, A. Rahimi, D. Rachkovskij, E. Osipov, and J. Rabaey, "Classification and recall with binary hyperdimensional computing: Tradeoffs in choice of density and mapping characteristics," *IEEE Transactions on Neural Networks and Learning Systems*, vol. 29, pp. 1–19, 04 2018.

[43] S. Bosch, A. S. de la Cerda, M. Imani, T. S. Rosing, and G. D. Micheli, "QubitHD: A stochastic acceleration method for hd computing-based machine learning," in

*arXiv:1911.12446*, 2019.

[44] A. Rahimi, S. Benatti, P. Kanerva, L. Benini, and J. M. Rabaey, "Hyperdimensional biosignal processing: A case study for emg-based hand gesture recognition," in *2016 IEEE International Conference on Rebooting Computing (ICRC)*, pp. 1–8, IEEE, 2016.

[45] M. Mitchell, *An introduction to Genetic Algorithms.* MIT Press, 1998.

[46] D. Bhandari, C. Murthy, and S. K. Pal, "Genetic algorithm with elitist model and its convergence," *International Journal of Pattern Recognition and Artificial Intelligence*, vol. 10, no. 06, pp. 731–747, 1996.

[47] P.-Y. Chen, H. Zhang, Y. Sharma, J. Yi, and C.-J. Hsieh, "Zoo: Zeroth order optimization based black-box attacks to deep neural networks without training substitute models," in *AISec*, 2017.

[48] I. J. Goodfellow, J. Shlens, and C. Szegedy, "Explaining and Harnessing Adversarial Examples," in *ICLR*, 2015.

[49] T. Hazan, G. Papandreou, and D. Tarlow, "Adversarial perturbations of deep neural networks," 2017.

[50] S.-J. Moon, V. Sekar, and M. K. Reiter, "Nomad: Mitigating arbitrary cloud side channels via provider-assisted migration," in *CCS*, 2015.

[51] T. Che, X. Liu, S. Li, Y. Ge, R. Zhang, C. Xiong, and Y. Bengio, "Deep verifier networks: Verification of deep discriminative models with deep generative models," *https://arxiv.org/abs/1911.07421*, 2019.

[52] N. Akhtar and A. Mian, "Threat of adversarial attacks on deep learning in computer vision: A survey," *IEEE Access*, vol. 6, pp. 14410–14430, 2018.

[53] N. Carlini and D. Wagner, "Towards evaluating the robustness of neural networks," in *S&P*, 2017.

[54] H. Xiao, B. Biggio, G. Brown, G. Fumera, C. Eckert, and F. Roli, "Is feature selection secure against training data poisoning?," in *International Conference on Machine Learning*, 2015.

[55] Z. Zhang, M. Liu, C. Zhang, Y. Zhang, Z. Li, Q. Li, H. Duan, and D. Sun, "Generating adversarial readable chinese texts," in *IJCAI*, 2020.

[56] Y. Cao, C. Xiao, B. Cyr, Y. Zhou, W. Park, S. Rampazzi, Q. A. Chen, K. Fu, and Z. M. Mao, "Adversarial sensor attack on lidar-based perception in autonomous driving," in *CCS*, 2019.

[57] N. Papernot, P. McDaniel, I. Goodfellow, S. Jha, Z. B. Celik, and A. Swami, "Practical black-box attacks against machine learning," in *AsiaCCS*, 2017.

[58] Y. Liu, X. Chen, C. Liu, and D. Song, "Delving into transferable adversarial examples and black-box attacks," in *ICLR*, 2017.

[59] L. Pengcheng, J. Yi, and L. Zhang, "Query-efficient black-box attack by active learning," in *ICDM*, 2018.

[60] X. Gao, Y.-a. Tan, H. Jiang, Q. Zhang, and X. Kuang, "Boosting targeted black-box attacks via ensemble substitute training and linear augmentation," *MDPI Applied Sciences*, vol. 9, no. 11, p. 2286, 2019.

[61] A. Ilyas, L. Engstrom, A. Athalye, and J. Lin, "Black-box adversarial attacks with limited queries and information," in *ICML*, 2018.

[62] C.-C. Tu, P. Ting, P.-Y. Chen, S. Liu, H. Zhang, J. Yi, C.-J. Hsieh, and S.-M. Cheng, "AutoZoom: Autoencoder-based zeroth order optimization method for attacking black-box neural networks," in *AAAI*, 2019.

[63] W. Brendel, J. Rauber, and M. Bethge, "Decision-Based Adversarial Attacks: Reliable Attacks Against Black-Box Machine Learning Models," in *ICLR*, 2018.

[64] N. Narodytska and S. Kasiviswanathan, "Simple black-box adversarial attacks on deep neural networks," in *CVPR Workshops*, 2017.

[65] W. Xu, Y. Qi, and D. Evans, "Automatically evading classifiers: A case study on PDF malware classifiers," in *NDSS*, 2016.

[66] B. Khaleghi, M. Imani, and T. Rosing, "Prive-HD: Privacy-preserved hyperdimensional computing," in *DAC*, 2020.